\documentclass[conference]{IEEEtran}
\IEEEoverridecommandlockouts
\pdfoutput=1
\usepackage[letterpaper, top=0.75in, bottom=1in, left=0.625in, right=0.625in]{geometry}
\usepackage{booktabs}

\usepackage{cite}
\usepackage{amsmath,amssymb,amsfonts}
\usepackage{algorithmic}
\usepackage{algorithm}
\usepackage{graphicx}
\usepackage{textcomp}
\usepackage{xcolor}
\usepackage{multirow}
\usepackage{comment}
\def\BibTeX{{\rm B\kern-.05em{\sc i\kern-.025em b}\kern-.08em
    T\kern-.1667em\lower.7ex\hbox{E}\kern-.125emX}}

\begin{document}

\title{Task-Adaptive Semantic Communications with Controllable Diffusion-based Data Regeneration\\
}

\author{\IEEEauthorblockN{Fupei Guo$^*$, Achintha Wijesinghe$^\dagger$,  Songyang Zhang$^*$, Zhi Ding$^\dagger$ }
\IEEEauthorblockA{$^*$University of Louisiana at Lafayette, Lafayette, LA, USA, 70504\\$^\dagger$University of California at Davis,  Davis, CA, USA, 95616 }}

\maketitle

\begin{abstract}
Semantic communications represent a new paradigm of next-generation networking that shifts bit-wise data delivery to conveying the semantic meanings for bandwidth efficiency. To effectively accommodate various potential downstream tasks at the receiver side, one should adaptively convey the most critical semantic information. This work presents a novel task-adaptive semantic communication framework based on diffusion models that is capable of dynamically adjusting the semantic message delivery according to various downstream tasks. 
Specifically, we initialize the transmission of a deep-compressed general semantic representation from the transmitter to enable diffusion-based coarse data reconstruction at the receiver. The receiver identifies the task-specific demands and generates textual prompts as feedback. Integrated with the attention mechanism, the transmitter updates the semantic transmission with more details to better align with the objectives of the intended receivers.
Our test results demonstrate the efficacy of the proposed method 
in adaptively preserving critical task-relevant information for semantic communications while preserving high compression efficiency.
\end{abstract}

\begin{IEEEkeywords}
Semantic communications, diffusion models, attention map, controllable data reconstruction.
\end{IEEEkeywords}

\section{Introduction}
The recent development of artificial intelligence (AI) has stimulated many novel explorations in wireless networking, such as artificial intelligence of things (AIoT) \cite{9264235} and vehicle-to-everything (V2X) communication \cite{8605302}, which require efficient data transmission with limited bandwidth. Traditional data-oriented communications focus on bit-wise data transmission \cite{8054694}, which can contain much redundant information for downstream tasks, suffering from heavy power and bandwidth overhead. However, in some scenarios, it is unnecessary to deliver exact raw data while conveying the key semantic message can adequately perform downstream tasks. For example, in autonomous driving, the shape of objects and distance to obstacles are more important than colors or patterns \cite{2024Diff-Go}. Thus, with the advancement of AI techniques, semantic communication has emerged as an important paradigm in next-generation communications, which prioritizes the delivery of meaningful information rather than simply transmitting raw bits \cite{9679803}. 

Semantic communications may manifest as task-oriented communications \cite{10183789}, where deep neural networks are used for semantic embedding and data recovery. One classical category of semantic communications uses 
the autoencoder (AE) structure. Established examples include joint semantic-channel coding \cite{10038754}, deep learning enabled semantic communications (DeepSC) \cite{9398576}, and masked vector quantized-variational autoencoder (VQ-VAE) \cite{10101778}. However, these AE-based approaches do not provide
strong interpretability of semantic embeddings and suffer from limited compression efficiency.  
To this end, more recent works in semantic communications focus on integrating generative learning models, where semantic representations condition the data regeneration. Notable generative AI-based semantic communications include generative semantic communications (GESCO) \cite{grassucci2023generative}, generative adversarial network (GAN) based frameworks \cite{lokumarambage2023wireless}, and token communications \cite{qiao2025token}. However, these approaches rely on the effective extraction of semantic representation, making semantic embedding a critical problem.

\begin{figure}
    \centering
    \includegraphics[width=0.92\linewidth]{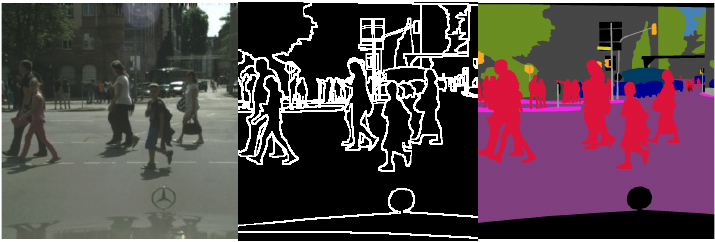}
    \vspace{-2mm}
    \caption{Examples of semantic representation for autonomous driving (from left to right): (a) original image, (b) edge map, and (c) segmentation. }
    \label{fig:illu}
    \vspace{-6mm}
\end{figure}

Existing semantic embedding methods are usually goal-oriented. For example, a local feedback block is introduced in Diff-GO \cite{2024Diff-Go}, which evaluates the quality of semantic representation at the transmitter, leading to additional computational burdens. As an extension, Diff-GO+ \cite{10947303}  offers multiple semantic representations. Undoubtedly, there is a tradeoff between performance and bandwidth usage for different semantic representations, where human decision may be involved. As shown in Fig. \ref{fig:illu}, the edge maps contain more details, while segmentation maps characterize abstract information of objects for autonomous driving. To enhance efficiency, \cite{wu2024semantic} introduces a semantic image transmission
system that allocates higher data-width to Regions of Interest
(ROI) when encoding an image.
However, these approaches assume a fixed downstream task at the receiver, 
leading to a prior-defined semantic representation which is usually unknown to the transmitter due to privacy. Moreover, real-world systems often involve dynamically changing tasks, making static semantic representation ineffective in achieving continuously high quality-of-services (QoS). Moreover, the re-transmission of an entirely new semantic is bandwidth-consuming, whereas a partial update with more details can be more effective. Thus, efficient adaptation of semantic message generation and transmission 
emerges as an urgent and exciting challenge.

\begin{figure*}[t]
    \centering
    \includegraphics[width=1\linewidth]{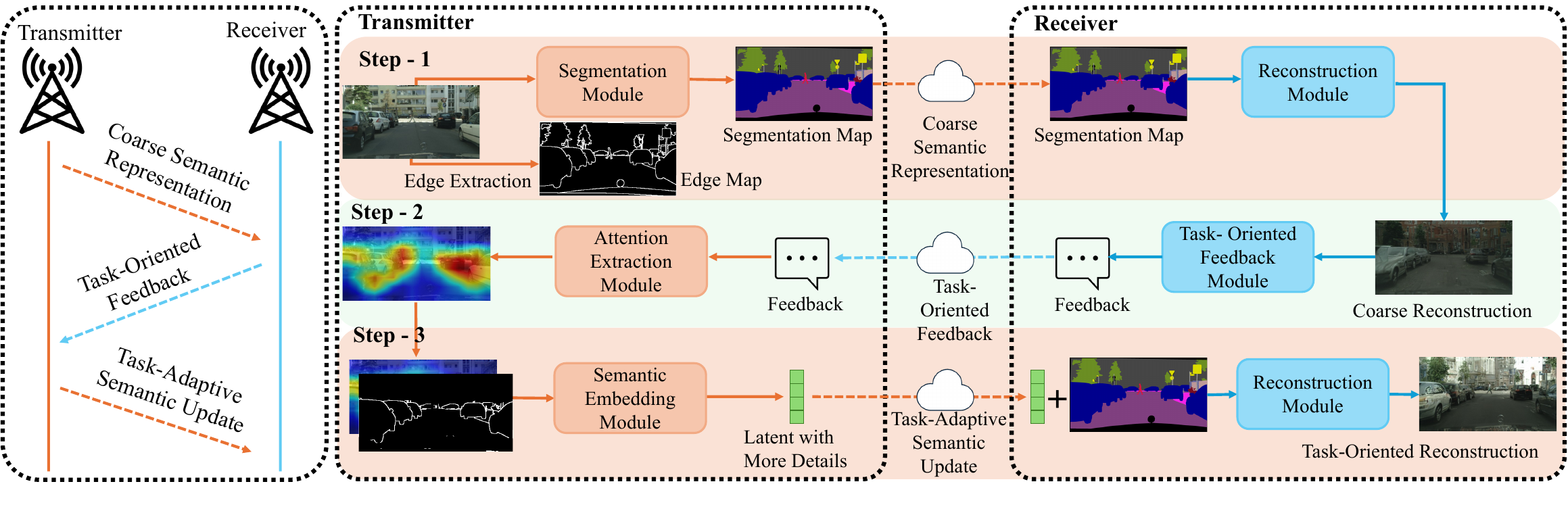}
    \vspace{-7mm}
    \caption{Diagram of the proposed three-phase semantic communication: 1) Step 1: the transmitter generates a general and coarse semantic representation to provide sketch information of raw media signal; 2) Step 2: the transmitter reconstructs the 
    original media signal via the reverse process of diffusion and also provide textual prompts as feedback; and 3) Step 3: transmitter updates the semantic representation with more task-oriented details, with which the receiver performs task-oriented reconstruction.}
    \label{fig:diagram}
    \vspace{-4mm}
\end{figure*}

To address the aforementioned problem, this work proposes a novel task-adaptive semantic communication framework based on conditional diffusion models, where semantic representation constrains data regeneration at the receiver. Specifically, we introduce a new three-phase interaction between transmitters and receivers, where a deep-compressed coarse semantic representation is initiated for the first interaction. Integrated with attention mechanisms and Contrastive Language–Image Pretraining (CLIP)–based semantic extraction, task-oriented feedback is sent back to the transmitter in the second interaction, after which the task-adaptive semantic latent with details is embedded for final data reconstruction. The proposed framework is capable of adjusting semantic representation flexibly to satisfy dynamically-changing downstream tasks. 
Our test results demonstrate the efficacy of the proposed method in adaptively preserving task-relevant information while keeping bandwidth efficiency high.

\section{Overall Architecture}
Before stepping into the details, we first briefly introduce our objective and overall structure of the proposed framework.

\subsection{Objective}
Departing from conventional semantic communications with known and fixed downstream tasks, this work focuses on scenarios in which the receiver may update their downstream tasks or semantic of interest while using the same 
received data. 
For example, to rescue trapped victims in inaccessible areas during disasters, the command center needs to explore different regions of the vision data collected by drones for victim searching. However, due to bandwidth limitations in emergency communications, transmitting entire high-resolution data is impossible. A more efficient way is to adaptively transmit the regions with a higher possibility of victim presence determined by the command center. 
Another typical example is real-time online diagnosis, where physicians can change their region of interest in imaging (e.g. MRI) data to identify different diseases during video transmission. Our objective is to adaptively adjust semantic representation and enable semantic transmission with ultra-high bandwidth efficiency according to the feedback from the receiver. In this work, we will use urban street images as an example, where downstream tasks are unknown to the transmitter, and the receiver may dynamically change their tasks, e.g., car detection and person detection. Related realistic applications include traffic monitoring and target tracking. Note that, our proposed framework can be easily generalized to other datasets and scenarios with multiple objectives in downstream tasks.

\subsection{Overall Structure}
The overall structure of the proposed framework is illustrated in Fig. \ref{fig:diagram}, which features a three-phase interaction between the transmitter and receiver for semantic adjustment. The general steps are illustrated as follows:
\subsubsection{Step 1} Since the transmitter is unaware of the current downstream task, it initializes a coarse semantic representation which can be deeply compressed and capture the general information. In this work, we consider the segmentation map for urban street scene, which is transmitted to the receiver for data regeneration. We also generate an edge map with more details but less compressible for semantic update in future steps, which is not transmitted in the first step.

\subsubsection{Step 2} At the receiver, a denoising diffusion probabilistic model (DDPM) \cite{ho2020denoising} is utilized for data regeneration, where the semantic representation serves as a condition to guide data generation. With the regenerated data, the receiver can determine whether the current transmission satisfies downstream tasks. If not, task-oriented textual feedback will be generated for the transmitter to update the semantic representation, which shall be bandwidth-efficient.

\subsubsection{Step 3} After receiving the feedback, an attention extraction module is deployed to generate the attention map to indicate the object of interest, after which a detailed edge map is masked by the attention map for a task-adaptive semantic update. By transmitting the updated semantic embedding to the receiver as a condition, new data shall be regenerated.

Here, the segmentation map offers high-level structural information, indicating the approximate locations of objects such as cars, people, and trees. The edge map provides fine-grained details relevant to the downstream task. By controlling the attention-masked edge map, the system can generate different task-oriented reconstructions, refining specific regions while maintaining overall scene consistency.

Note that, if the receiver changes its downstream task and semantic of interest, Steps 2-3 will be repeated for task-adaptive semantic updating and data regeneration.

\section{Method}
We now present the detailed design of each module at the transmitter and receiver sides.
\subsection{Modules at Transmitter}

The transmitter is charged with generating a suitable semantic representation. It consists of three key modules: 1) semantic generation to generate a segmentation map in Step 1; 2) attention extraction to translate receiver feedback from text to attention map in Step 2, which characterizes the region of interest (ROI) automatically; 3) semantic embedding for representation update with task-adaptive details in Step 3.

\subsubsection{Semantic Generation (in Step-1)} 
This work views segmentation maps as coarse semantic representation to initiate the three-phase process. 
Consider an input image $\mathbf{x} \in \mathbb{R}^{H \times W \times C}$, where $H$, $W$, and $C$ represent the height, the width, and the number of color channels, respectively. Its semantic segmentation map $\mathbf{x}_{\text{seg}} \in \mathbb{R}^{H \times W \times K}$ can be extracted using a pre-trained segmentation neural network, such as Unet and Residual Networks (ResNet) \cite{chen2017rethinking}, where $K$ denotes the number of semantic categories. In this work, ResNet and one-hot encoding are applied for segmentation extraction and compression, respectively.
Since we aim to adjust the semantic representation with more details in the ROI after receiving feedback, an edge map $\mathbf{x}_{\text{edge}} \in \mathbb{R}^{H \times W}$ is also generated for future updating in Step 3. Canny-based learning approach is utilized for edge extraction based on the instance map \cite{6885761}.

\subsubsection{Attention Extraction (in Step-2)} \label{sec:att}
The Attention Extraction Module aims to generate task-relevant attention maps based on the feedback sent by receivers capable of identifying the most critical part in the received image.

\begin{figure}[htb]
    \centering
    \vspace{-2mm}
    \includegraphics[width=1\linewidth]{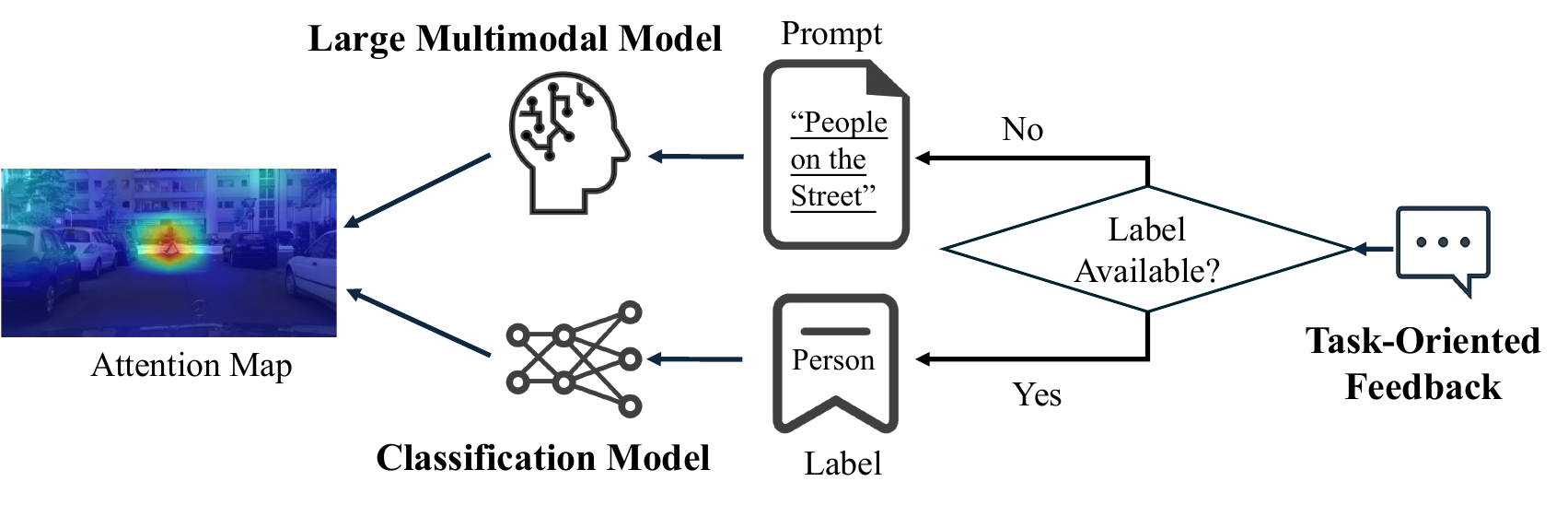}
    \vspace{-8mm}
    \caption{Two types of attention schemes: 1) class activation map for classification tasks; and 2) CLIP-based attention maps for general tasks.}
    \label{fig:Attention extraction module}
    \vspace{-2mm}
\end{figure}

To ensure bandwidth efficiency, we mainly consider two types of feedback: 1) class labels that correspond to known object categories (e.g., ``car'', ``person''); or (2) free-form textual prompts (e.g., ``I am interested in people on the street.'') that describe objectives more flexibly, for which different attention mechanisms are investigated, as shown in Fig.~\ref{fig:Attention extraction module}. Here, we describe the details of each method.

For feedback based on known classes, a pre-trained classifier (e.g., ResNet-50) is used to generate class activation maps (CAMs)~\cite{zhou2016learning}, highlighting pixel importance for the given label. Given an image $\mathbf{x}$ and a target class label $c$, the classification neural network produces a set of deep convolutional features $\mathbf{F} \in \mathbb{R}^{H' \times W' \times L}$ from the final convolutional layer, where $L$ is the number of channels. A class-specific attention map $\mathbf{A}_F$ can be computed by a weighted sum over all channels, i.e.,
\begin{equation}
    \mathbf{A}_F = \sum_{k=1}^{L} w_{c,k} \cdot \mathbf{F}_k,
\end{equation}
using $w_{c,k}$ as the classification weight for class $c$ of $k$-th channel in the final fully connected layer.
Note that $\mathbf{F}_k$ is the $k$-th channel of $\mathbf{F}$. Next, $\mathbf{A}_F$ is upsampled to the original image resolution, i.e., $\mathbf{A}_C\in\mathbb{R}^{H\times W}$, and is normalized to capture the spatial attention distribution.

If the receiver sends a textual description as feedback, the CLIP Attention Extraction \cite{radford2021learning} generates the corresponding attention map. This approach is particularly effective when the target object class is not given or when the objective is described using natural languages. Given an input image $\mathbf{x} \in \mathbb{R}^{H \times W \times C}$ and a set of received textual descriptions $\mathcal{T} = \{ t_1, t_2, ..., t_n \}$, the attention map $\mathbf{A}_{\text{CLIP}}$ is computed as
\begin{equation}
    \mathbf{A}_{\text{CLIP}} = f_{\text{CLIP}}(\mathbf{x}, \mathcal{T}),
\end{equation}
where $f_{\text{CLIP}}(\cdot, \cdot)$ measures the similarity between spatial image features and text embeddings. The resulting attention map $\mathbf{A}_{\text{CLIP}} \in \mathbb{R}^{H \times W}$ highlights regions in the image that are most relevant to the given textual description.

By supporting both label-based and text-based feedbacks, Attention Extraction Module enables transmitters to flexibly adapt to a wide range of downstream tasks, facilitating controllable semantic characterization in real-world scenarios.

\subsubsection{Semantic Embedding (in Step-3)} In this step, the semantic representation is updated with more details in the ROI of the image, particularly for those with higher energy in the attention map. Specifically, part of the previously derived edge map is transmitted. Instead of transmitting the entire edge map, a mask $\mathbf{M}_A$ is generated from attention map using a predefined threshold $\tau$, i.e., $\mathbf{M}_A=\mathbb{I}(\mathbf{A}>\tau)$, where $\mathbb{I}(\cdot)$ is the indicator function to generate the binary mask. Next, the masked edge map with task-specific attention is defined as 
\begin{equation}
    \mathbf{x}_{\text{att}} = \mathbf{M}_A \odot \mathbf{x}_{\text{edge}},
\end{equation}
where $\odot$ denotes the Hadamard product. This process preserves only the most relevant edges, effectively reducing redundant information.

To further enhance spectrum efficiency, the task-specific edge representation $\mathbf{x}_{\text{att}}$ is partitioned into $P$ nonoverlapping patches of size $n \times n$, denoted by $\{\mathbf{p}_1, \mathbf{p}_2, \ldots, \mathbf{p}_P\}$. All zero patches are discarded. The remaining informative patches are flattened and concatenated with location information to form a latent feature vector $\mathbf{z}_{\text{edge}}$ for transmission, i.e.,
\begin{equation}
    \mathbf{z}_{\text{edge}} = \text{Concat}\left( \text{Flatten}(\mathbf{p}_i) \ \big| \ \sum_{j} p_{i,j} > 0 \right).
\end{equation}

Note that, our proposed framework can be integrated with conventional image coding techniques, such as ``JPEG" and ``PNG" for further compression.

\subsection{Modules at Receiver}
The receiver focuses on generating data based on semantic representation and provides feedback to the transmitter, using the detailed structure introduced below.

\subsubsection{Reconstruction Module (in Step-1\& Step-3)} To generate high-fidelity images based on semantic representation, a denoising diffusion probabilistic model (DDPM) is deployed.

Given the segmentation map $\mathbf{x}_{\text{seg}}$ and the attention-masked edge map $\mathbf{x}_{\text{att}}$, the diffusion model reconstructs the image by learning the reverse process of a Gaussian noise injection. The reconstruction follows a probabilistic framework that maximizes the likelihood $p_{\theta}(\mathbf{x}_0|\mathbf{x}_{\text{seg}}, \mathbf{x}_{\text{att}})$, which can be modeled as a Markov chain with Gaussian transitions, initialized with a prior distribution $\mathbf{x}_T \sim \mathcal{N}(\mathbf{0}, \mathbf{I})$, formulated by
\begin{equation}
    p_{\theta}(\mathbf{x}_{0:T}|\mathbf{x}_{\text{seg}}, \mathbf{x}_{\text{att}}) =
    p(\mathbf{x}_T) \prod_{t=1}^{T} p_{\theta}(\mathbf{x}_{t-1} | \mathbf{x}_t, \mathbf{x}_{\text{seg}}, \mathbf{x}_{\text{att}}).
\end{equation}
Here, each transition is parameterized as
    \begin{align}
        &p_{\theta}(\mathbf{x}_{t-1} | \mathbf{x}_t, \mathbf{x}_{\text{seg}}, \mathbf{x}_{\text{att}}) 
       \\ &= \mathcal{N} \Big( \mathbf{x}_{t-1};
        \mu_{\theta}(\mathbf{x}_t,\mathbf{x}_{\text{seg}}, \mathbf{x}_{\text{att}}, t), 
          \sigma_{\theta}(\mathbf{x}_t, \mathbf{x}_{\text{seg}}, \mathbf{x}_{\text{att}}, t) \Big).
    \end{align}

The forward process $q(\mathbf{x}_{1:T}|\mathbf{x}_0)$ gradually adds Gaussian noise to the input according to a predefined variance schedule $\beta_1, ..., \beta_T$. Using the notation $\alpha_t = \prod_{s=1}^{t} (1 - \beta_s)$, the forward process is defined as
\begin{equation}
    q(\mathbf{x}_t | \mathbf{x}_0) = \mathcal{N}(\mathbf{x}_t; \sqrt{\alpha_t} \mathbf{x}_0, (1 - \alpha_t) \mathbf{I}).
\end{equation}

\begin{figure}[t]
    \centering
 \includegraphics[width=0.8\linewidth]{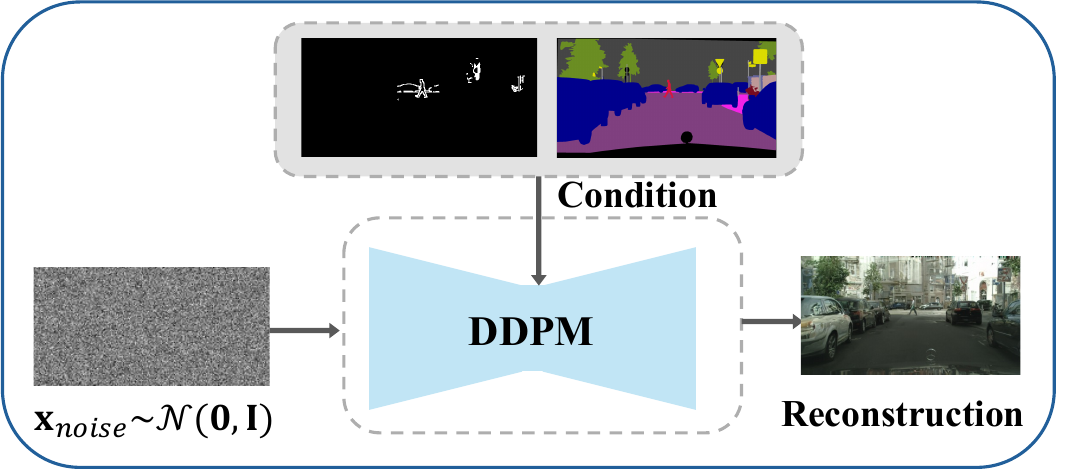}
 \vspace{-2mm}
    \caption{Structure of the diffusion-based reconstruction module.}
    \label{fig:reconstruction}
    \vspace{-3mm}
\end{figure}

In Step-1, the edge map is all-zero since no preference is provided. In Step-3, the masked edge map $\mathbf{x}_{\text{att}}$ is decoded from the received latent $\mathbf{z}_{\text{edge}}$
This conditional DDPM framework enables adaptive semantic reconstruction, allowing the receiver to request different levels of details based on task requirements. The structure of our reconstruction module is illustrated in Fig.~\ref{fig:reconstruction}. 

\subsubsection{Task-Oriented Feedback Module (in Step 2)}
In the three-phase interaction, the receiver plays an active role in providing feedback to guide semantic refinement in Step-3. As mentioned in Section \ref{sec:att}, we consider two types of feedback: 1) labels; and 2) textual prompts. When the downstream task is aligned with known object categories, the receiver can return a discrete class label such as ``car" or ``person". This class-level feedback is efficient and well-suited for clearly defined objectives. Meanwhile, when the target is more abstract or difficult to express using defined categories, the receiver may return a textual prompt, such as ``people walking on the sidewalk". These prompts can be manually defined or automatically generated by a multimodal large language model, enabling the system to handle open-ended or previously unseen task descriptions.

\section{Experiments}

\subsection{Experimental Setup}

\textbf{Dataset.} We evaluate on the Cityscapes dataset~\cite{Cordts2016Cityscapes}, which provides pixel-level annotations for 35 classes across diverse urban scenes. All the images are resized to $256 \times 512$ to reduce the computation cost. The dataset is split into 70\% for training, 20\% for validation, and 10\% for testing.

\textbf{Hyperparameters.} The diffusion model uses a U-Net with 5 downsampling/upsampling stages and channel sizes from 256 to 1024. Attention blocks are applied at the lowest-resolution levels, including the bottleneck, to enhance spatial awareness during reconstruction.
CLIP features are extracted using the pre-trained model from~\cite{radford2021learning}. 
The attention map is derived from ResNet-50 \cite{chen2017rethinking}.

 \begin{table}[t]
    \centering
    \caption{Comparison of different methods using LPIPS and FID.}
    \label{tab:comparison with baseline}
    \begin{tabular}{lcc}
        \toprule
        Method & LPIPS $\downarrow$ & FID $\downarrow$ \\
        \midrule
        SPADE \cite{park2019semantic} & 0.546 & 103.240 \\
        CC-FPSE \cite{liu2019learning} & 0.546 & 245.900 \\
        SMIS \cite{zhu2020semantically} & 0.546 & 87.580 \\
        OASIS \cite{sushko2020you} & 0.561 & 104.030 \\
        SDM \cite{wang2022semantic} & 0.549 & 98.990 \\
        GESCO\cite{grassucci2023generative} & 0.433 & 102.22 \\
        Diff-GO (Segmentation) \cite{2024Diff-Go} &  0.414 & 96.929 \\
        Ours (Car-CAM-Attn) & 0.410 & 91.450 \\
        Ours (Person-CAM-Attn) & 0.410 & 94.690 \\
        Ours (Car-Person-CAM-Attn) & 0.407 & 89.788 \\
        Ours (Car-CLIP-Attn) & 0.400 & 87.980 \\
        Ours (Person-CLIP-Attn) & 0.400 & 90.770 \\
        Ours (Car-Person-CLIP-Attn) & 0.400 & 88.906 \\
        Ours (All-Attn) & 0.394 & 88.920 \\
        \bottomrule
    \end{tabular}
    \vspace{-3mm}
\end{table}
\begin{table*}[t]
    \centering
    \renewcommand{\arraystretch}{1.3}
    \caption{Comparison of task performance: object detection (MSE$\downarrow$, mIoU$\uparrow$) and depth estimation (RMSE$\downarrow$, SI-RMSE$\downarrow$).}
    \vspace{-2mm}
    \label{tab:combined_task_performance}
    \begin{tabular}{lcc|cc|c|c|c}
        \toprule
        \multirow{2}{*}{Semantic Representation} 
        & \multicolumn{2}{c|}{MSE$\downarrow$ (Object Counting)}  
        & \multicolumn{2}{c|}{mIoU$\uparrow$ (Object Detection)} 
        & \multirow{2}{*}{RMSE$\downarrow$ (Depth)} 
        & \multirow{2}{*}{SI-RMSE$\downarrow$ (Depth)} 
        & \multirow{2}{*}{Compression Rate} \\
        \cmidrule(lr){2-3} \cmidrule(lr){4-5}
        & Person  & Car & Person & Car& & & \\
        \midrule
           No-Attn (Step-1 Only) & 4.954 & 5.980 & 74.43 & 82.37 & 0.0631 & 0.2780 & 31.22  \\
         All-Attn & \textbf{1.145} & 1.360 & 81.06 & \textbf{85.91} & 0.0609 & \textbf{0.2420} & 15.46 \\
         \midrule
        Car-CAM-Attn & 2.906 & \underline{1.268} & 76.33 & 85.21 & 0.0598 & 0.2563 & 20.50 \\
        Person-CAM-Attn & 1.513 & 4.466 & \underline{\textbf{81.55}} & 81.48 & 0.0593 & 0.2592 & 22.71 \\
        Car-Person-CAM-Attn & \underline{1.206} & {1.391} & 80.82 & \underline{85.75} & 0.0596 & 0.2629 & 19.09 \\
        \midrule
        Car-CLIP-Attn & 1.757 & 2.669 & 78.64 & 85.43 & 0.0609 & 0.2628 & 21.03 \\
        Person-CLIP-Attn & 1.585 & 1.780 & \underline{81.10} & 83.69 & 0.0594 & 0.2597 & 19.97 \\
        Car-Person-CLIP-Attn & \underline{1.391} & \underline{\textbf{1.206}} & 80.82 & \underline{85.75} & \textbf{0.0588} & 0.2595 & 18.00 \\
        \bottomrule
    \end{tabular}
\end{table*}
\begin{figure*}[t]
    \centering
    \includegraphics[width=0.95\linewidth]{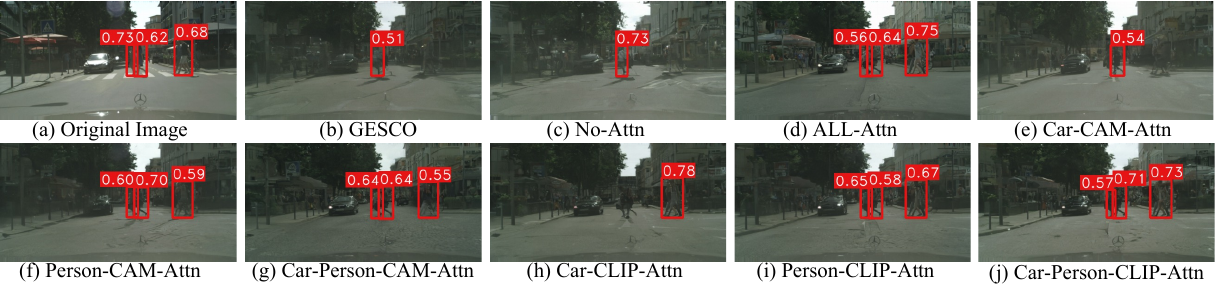}
    \vspace{-4mm}
    \caption{Object Detection: Results of Reconstructed Images under Different Conditions: people bounded by blue boxes.}
\label{fig:reconstruction_detection}
\vspace{-2mm}
\end{figure*}
\subsection{Evaluation of Reconstruction Quality}
To assess the quality of reconstructed images, we compare our framework with both semantic image synthesis models (e.g., SPADE~\cite{park2019semantic}, CC-FPSE~\cite{liu2019learning}, SMIS~\cite{zhu2020semantically}, OASIS~\cite{sushko2020you}, and SDM~\cite{wang2022semantic}), and diffusion-based semantic communications (e.g., Diff-GO\cite{2024Diff-Go} and GESCO\cite{grassucci2023generative}). Parts of the baseline results are sourced from GESCO~\cite{grassucci2023generative}. In our approaches, we test different combinations of attentions for ``cars" and ``person" as exemplary objects of interest.
All the diffusion-based results are re-trained under the same setup for fair comparison. 
To measure the perceptual quality of data reconstruction, we employ Fréchet Inception Distance (FID) and Learned Perceptual Image Patch Similarity (LPIPS). Lower FID and LPIPS scores indicate better performance.


The results are presented in Table~\ref{tab:comparison with baseline}, showing that semantic communications based on generative AI outperform the conventional image synthesis model. Since our proposed methods enable the transmission of more task-specific details with high bandwidth efficiency, all our methods achieve superior performance over
the GESCO and Diff-GO benchmarks. Since Table~\ref{tab:comparison with baseline} focuses on the overall image regeneration, CLIP-based attentions have better performance than CAM-based attentions, due to the generalization ability of CLIP.
Note that, our proposed three-phase interaction semantic updating can be flexibly integrated with Diff-GO and GESCO to facilitate task-adaptive adjustment for data reconstruction.

\subsection{Performance on Downstream Tasks}

To evaluate the effectiveness of our method in practical scenarios, two applications are considered: 1) object detection with adjustable interests; and 2) depth estimation. 

\subsubsection{Object Detection}
Our first task focuses on object detection, where receivers may dynamically change their interest in ``cars" and ``person". Thus, we implement both car detection and person detection to mimic dynamically changing downstream tasks. Note that, there are more than two types of objects in all images. Here, a pre-trained YOLOv8~\cite{reis2024realtimeflyingobjectdetection} model is used for different object detection.

To evaluate the performance, we consider two metrics: 1) number of target objects; and 2) localization accuracy. For the number of target objects, we count the detection with confidence scores above a certain threshold (e.g., 0.5 in this work). Mean squared error (MSE) is calculated between the number of predicted objects \( \hat{N}_{C_i} \) and the corresponding ground truth count \( {N}_{C_i} \) for the $i$-th image, denoted by
\(\text{MSE} = \frac{1}{K} \sum_{i=1}^{K} (\hat{N}_{C_i} - {N}_{C_i})^2,\)
where \( K \) is the total number of evaluated images. Lower MSE indicates better consistency in the presence of objects. 

To assess localization accuracy, the mean Intersection over Union (mIoU$\uparrow$) is computed as 
$\text{mIoU} = \frac{1}{K} \sum_{i=1}^{K} \frac{|\hat{b}_i \cap b_i|}{|\hat{b}_i \cup b_i|}$, 
where $\hat{b}_i$ and $b_i$ denote the predicted and ground-truth bounding boxes for the $i$-th object, and $K$ is the total number of evaluated instances. Moreover, the corresponding compression rate is also provided, where $\text{CR}$ is the
ratio of full data size to the compressed data size. Here, PNG is integrated with our proposed encoding scheme to further compress the data.

The results are presented in Table \ref{tab:combined_task_performance}, where the best performance is highlighted in bold and underlined for all methods and each attention mechanism, respectively. From the results, all-attention with the entire edge map achieves the best overall performance but lowest compression ratio since it attempts to deliver all the details. No-Attn scheme achieves the best compression ratio at sacrifice of performance. If a specific feedback is available, the performance of detection for the target object significantly increases with better compression ratio. Since CLIP attention is generated from textual language and artifacts can be introduced, CAM-based approaches have more task-oriented improvements. Another interesting observation is that, although All-Att semantic achieves good overall performance, it does not always perform the best for all individual tasks, which is due to the redundant information in the edge map. The visualization results are shown in Fig. \ref{fig:reconstruction_detection}, where GESCO fails to detect three instances of ``person" while our methods successfully complete the tasks.
In summary, our proposed method demonstrates superior performance in the detection of the target object with a higher compression ratio, which validates its effectiveness in semantic adjustment.

\subsubsection{Depth Map Estimation}
We also evaluate the accuracy of depth estimation, where 
we report both the standard Root Mean Squared Error (RMSE$\downarrow$) and the scale-invariant RMSE (SI-RMSE$\downarrow$).
As presented in Table~\ref{tab:combined_task_performance}, the task-adaptive semantic representation achieves similar performance to all-attention sharing, with better compression rate and bandwidth efficiency, which validate the effectiveness of our proposed method in general downstream tasks.

\section{Conclusion}
This work proposes a task-adaptive semantic communication framework that features a three-phase process for controllable diffusion-based data reconstruction.
By introducing task-oriented conditioning, the system enables the receiver to refine the semantic of interest according to the current downstream tasks. Integrated
with attention mechanisms and CLIP-based semantic extraction, the transmitter can effectively identify the receiver's requirement and update a task-adaptive semantic representation with high bandwidth efficiency. Test results demonstrate that our approach can provide a good trade-off between compression rate and task-oriented reconstruction quality. Our proposed framework can be flexibly adopted to many semantic communication frameworks based on generative AI, offering a promising mechanism for future design of spectrum-efficient communications.


\bibliographystyle{IEEEtran} 
\bibliography{reference}

\end{document}